\documentclass[runningheads]{llncs}

\usepackage[T1]{fontenc}
\def\doi#1{\href{https://doi.org/\detokenize{#1}}{\url{https://doi.org/\detokenize{#1}}}}
\usepackage{graphicx}
\usepackage{microtype}
\usepackage{amsmath}
\usepackage{amsfonts}
\usepackage{microtype}
\usepackage{graphicx}
\usepackage{multirow}
\usepackage{booktabs}
\usepackage{hhline}
\usepackage{xspace}
\usepackage{color}
\usepackage{enumitem}
\usepackage{amsmath}
\usepackage{amssymb}
\usepackage{pifont}
\usepackage{amsfonts}
\usepackage{microtype}
\usepackage{graphicx}
\usepackage{multirow}
\usepackage{booktabs}
\usepackage{hhline}
\usepackage{xspace}
\usepackage{color}
\usepackage{enumitem}
\usepackage{ulem}
\usepackage{comment}
\usepackage{bbding}
\usepackage[pagebackref=true,breaklinks=true,colorlinks=true,bookmarks=false,citecolor=blue,urlcolor=black,linkcolor=blue]{hyperref}
\newcommand{\cmark}{\ding{51}}%
\newcommand{\xmark}{\ding{55}}%
\usepackage{listings}
\begin{document}
\title{Multi-Modal Masked Autoencoders for Medical Vision-and-Language Pre-Training}
\titlerunning{Multi-Modal Masked Autoencoders}
\author{
    Zhihong Chen  \inst{1} \and 
    Yuhao Du\inst{1} \and       
    Jinpeng Hu\inst{1} \and     
    Yang Liu\inst{1} \and \\    
    Guanbin Li\inst{2}\thanks{Corresponding authors are Guanbin Li and Xiang Wan.} \and 
    Xiang Wan\inst{1,3*} \and   
    Tsung-Hui Chang\inst{1}     
    }
\authorrunning{Z. Chen et al.}
\institute{Shenzhen Research Institute of Big Data \\
    The Chinese University of Hong Kong, Shenzhen, China \\ \and
    Sun Yat-sen University, Guangzhou, China \\ \and
    Pazhou Lab, Guangzhou, China \\
    \email{liguanbin@mail.sysu.edu.cn} \\
    \email{wanxiang@sribd.cn}}

\maketitle

\begin{abstract}
Medical vision-and-language pre-training provides a feasible solution to extract effective vision-and-language representations from medical images and texts.
However, few studies have been dedicated to this field to facilitate medical vision-and-language understanding.
In this paper, we propose a self-supervised learning paradigm with multi-modal masked autoencoders (M$^3$AE), which learn cross-modal domain knowledge by reconstructing missing pixels and tokens from randomly masked images and texts.
There are three key designs to make this simple approach work.
First, considering the different information densities of vision and language, we adopt different masking ratios for the input image and text, where a considerably larger masking ratio is used for images.
Second, we use visual and textual features from different layers to perform the reconstruction to deal with different levels of abstraction in visual and language.
Third, we develop different designs for vision and language decoders (i.e., a Transformer for vision and a multi-layer perceptron for language).
To perform a comprehensive evaluation and facilitate further research, we construct a medical vision-and-language benchmark including three tasks.
Experimental results demonstrate the effectiveness of our approach, where state-of-the-art results are achieved on all downstream tasks.
Besides, we conduct further analysis to better verify the effectiveness of different components of our approach and various settings of pre-training. The source code is available at~\url{https://github.com/zhjohnchan/M3AE}.

\end{abstract}

\section{Introduction}
Medical data is inherently multi-modal, where vision (e.g., radiography, magnetic resonance imaging, and computed tomography) and language (e.g., radiology reports and medical texts) are two primary modalities.
Medical vision-and-language pre-training (Med-VLP) aims to learn generic representations from large-scale medical image-text data, which can be transferred to various medical vision-and-language tasks (e.g., medical visual question answering (Med-VQA), medical image-text classification, and medical image-text retrieval).
It is an essential technique for jointly understanding medical images and texts, which can be challenging due to the lack of large-scale labeled data and domain knowledge.

Although vision-and-language pre-training (VLP) has drawn sustaining attention \cite{chen2020uniter,huang2020pixel,kim2021vilt,su2019vlbert,tan2019lxmert}, there are only several studies on VLP in the medical domain.
\cite{li2020comparison} directly applied four VLP models (i.e., LXMERT \cite{tan2019lxmert}, VisualBERT \cite{li2019visualbert}, UNITER \cite{chen2020uniter}, and PixelBERT \cite{huang2020pixel}) to a medical image-text classification task yet found that these models did not perform as well as they did in the general domain when there is no domain-specific information integrated.
Therefore, \cite{khare2021mmbert} proposed to perform the pre-training on medical image-text pairs to capture medical knowledge, but their evaluation was conducted only on Med-VQA despite the promising improvement is observed.
The most related work to ours is \cite{moon2021multi}, which pre-trained a Med-VLP model and verified its effectiveness on various downstream tasks.
Yet it is limited to the chest X-ray, and more importantly, the pre-training was not performed in a self-supervised manner (i.e., using the diagnosis labels).
Moreover, previous studies all used convolutional neural networks (CNNs) as their visual backbones, which limited their simplicity and effectiveness, and purely Transformer-based models \cite{vaswani2017transformer} are not exploited.
Therefore, it is essential to design an appropriate Med-VLP approach from four perspectives, including data (e.g., pre-training corpus), models (e.g., purely Transformer-based models), objectives (e.g., more suitable pre-training objectives), and evaluation (e.g., designs of the downstream benchmark) to promote Med-VLP.

In this paper, we propose an effective yet simple approach to Med-VLP by a multi-modal masked autoencoder (M$^{3}$AE) based on purely Transformer-based models.
Our M$^{3}$AE masks random patches of the input image and random tokens of the input text and reconstructs the missing pixels and tokens.
We develop the designs of M$^{3}$AE from three perspectives to make this simple approach work:
(i) It uses different masking ratios for the input images and texts owing to different information densities of vision and language;
(ii) It selects visual and textual features from distinct layers to perform the construction considering the different levels of abstraction in vision and language;
(iii) It has two different decoder designs for vision and language, where a Transformer model and a multi-layer perceptron (MLP) are used for vision and language decoding, respectively.
As a result, the proposed method is able to learn cross-modal domain-speciﬁc knowledge from large-scale medical image-text datasets in a self-supervised manner and does not require fine-grained annotations on either images or texts, resulting in better applicability.
We perform the pre-training on two large-scale medical image-text datasets, i.e., ROCO \cite{pelka2018roco} and MedICaT \cite{subramanian2020medicat}.
To verify the effectiveness of our approach and facilitate further research, we construct a medical vision-and-language understanding benchmark including three tasks (i.e., Med-VQA, medical image-text classification, and medical image-text retrieval).
Experimental results show that our approach outperforms previous studies on all downstream tasks.
In addition, several analyses are also performed to analyze the effectiveness of different components and various settings of pre-training.

\section{The Proposed Approach}
\label{sec:approach}
\begin{figure*}[t]
\centering
\includegraphics[width=0.98\textwidth, trim=0 0 0 0]{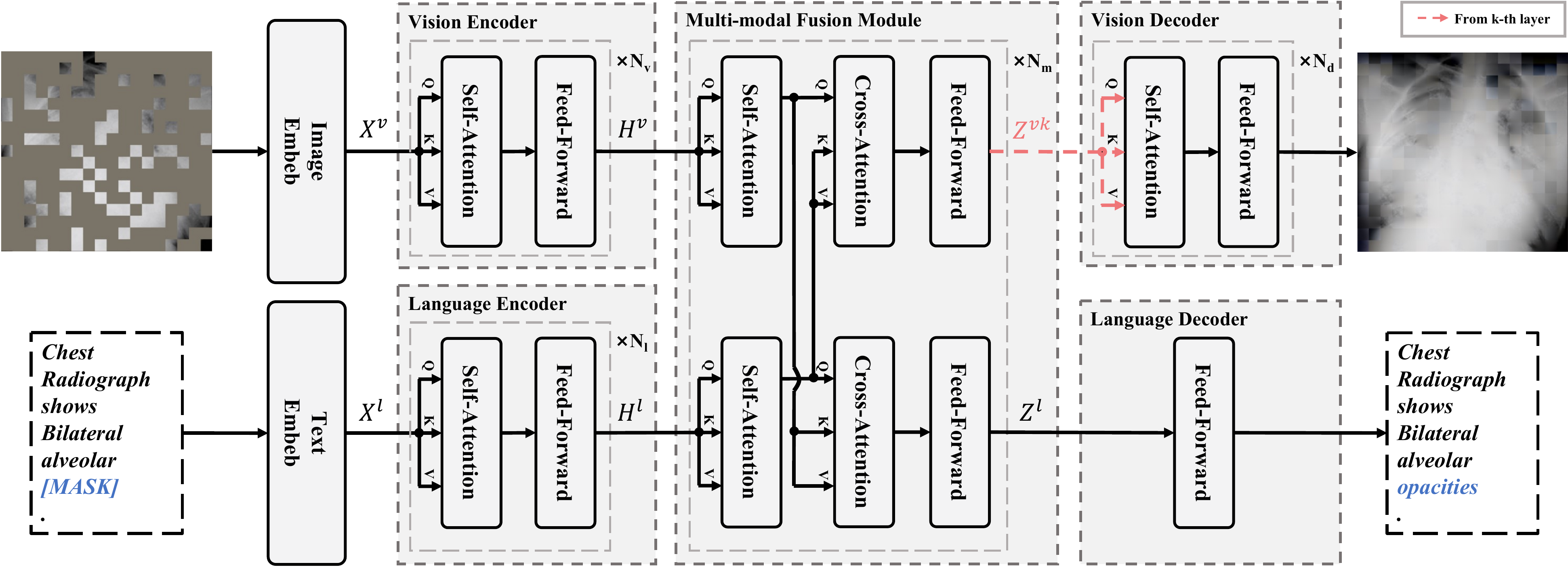}
\caption{The overall architecture of our proposed approach, where the vision encoder, language encoder, multi-modal fusion module, and decoders are shown in dash boxes.}
\label{fig:framework}
\end{figure*}
We adopt the pre-train-and-fine-tune paradigm for medical vision-and-language understanding.
In the pre-training stage, the framework develops a variety of pretext tasks to train the model using medical image-text pairs.
Formally, given a medical image $I$ and its corresponding description text $T$, the model is trained to minimize the objective through
\begin{equation}
    \theta^{*},\theta_{1}^{*},...,\theta_{S}^{*} = \mathop{\arg\min}_{\theta,\theta_{1},...,\theta_{S}}
    \sum_{s=1}^{S} L_{s}
    (Y_{s}, \mathcal{D}_{\theta_{s}}(\mathcal{M}_{\theta}(I, T))
\end{equation}
where $S$ is the number of pretext tasks, $L_{s}$ are the loss functions of pretext tasks, $\mathcal{D}_{\theta_{s}}$ are the decoders with their parameters $\theta_{1},...,\theta_{S}$, and $\mathcal{M}_{\theta}$ is the backbone model with its parameters $\theta$.
An overview of the proposed approach is demonstrated in Fig. \ref{fig:framework}, where the details of the backbone model architecture and multi-modal masked autoencoders are illustrated in the following subsections.

\subsection{The Backbone Model Architecture}
Our backbone model can be partitioned into three major components, i.e., the vision encoder, language encoder, and multi-modal fusion module.

\noindent\textbf{Vision Encoder}~
In this paper, we focus on purely Transformer-based models and study the use of vision Transformer (ViT) for the vision encoder.
In ViT, an image $I \in \mathbb{R}^{H \times W \times C}$ is first segmented into patches $\{p_{1}, p_{2}, ..., p_{N}\}$, where $H \times W$ is the image resolution, $C$ is the number of channels, $p_{n} \in \mathbb{R}^{P^{2} \times C}$ and $P \times P$ is the patch resolution.
Then the patches are flattened and linearly projected into patch embeddings through a linear transformation $E^{v} \in \mathbb{R}^{P^{2}C \times D}$ with a special learnable token embedding $p_{I} \in \mathbb{R}^{D}$ prepended for the aggregation of visual information.
Therefore, the input representations are obtained via summing up the patch embeddings, learnable 1D position embeddings $E^{v}_{pos} \in \mathbb{R}^{(N+1) \times D}$:
\begin{equation}
    X^{v} = [p_{I}; p_{1}E^{v}; p_{2}E^{v}; ...; p_{N}E^{v}] + E^{v}_{pos}
\end{equation}
Finally, $X^{v}$ is fed into a transformer model with $N_{v}$ Transformer layers to obtain the contextualized image representations $ H^{v} = [h^{v}_{I}; h^{v}_{1}; h^{v}_{2}; ...; h^{v}_{N}]$.

\noindent\textbf{Language Encoder}~
In the language encoder, we follow BERT \cite{devlin2019bert} to tokenize the input text to subword tokens $\{w_{1},w_{2},...,w_{M}\}$ by WordPiece \cite{wu2016google}, where the tokens $w_{m} \in \mathbb{R}^{V}$ are represented in one-hot form and $V$ is the vocabulary size.
Then the tokens are linearly projected into embeddings through a linear transformation $E^{l} \in \mathbb{R}^{V\times D}$. 
Afterwards, a start-of-sequence token embedding $w_{T} \in \mathbb{R}^{D}$ and a special boundary token embedding $w_{SEP} \in \mathbb{R}^{D}$ are added to the text sequence.
Therefore, the text input representations are computed via summing up the token embeddings and text position embeddings $E^{l}_{pos} \in \mathbb{R}^{(M+2) \times D}$:
\begin{equation}
    X^{l} = [w_{T}; w_{1}E^{l}; ...; w_{M}E^{l}; {w}_{SEP}] + E^{l}_{pos}
\end{equation}
Finally, $X^{l}$ is fed into a transformer model with $N_{l}$ Transformer layers to obtain the contextualized text representations $ H^{l} = [h^{l}_{T}; h^{l}_{1}; h^{l}_{2}; ...; h^{l}_{M}; h^{l}_{SEP}]$.

\noindent\textbf{Multi-modal Fusion Module}~
We adopt the co-attention mechanism in the multi-modal fusion module to fuse the contextualized representations from images and texts.
In detail, the multi-modal fusion module consists of two Transformer models, each of which is a stack of $N_{m}$ Transformer layers.
In each Transformer layer, there are three sub-layers, i.e., a self-attention sub-layer, a cross-attention sub-layer, and a feedforward sub-layer.
The attention mechanism is applied in the self-attention and cross-attention sub-layers and it is defined as
\begin{equation}
    \text{ATTN}(Q, K, V)=\operatorname{softmax}\left(Q K^{\top}\right) \cdot V
\end{equation}
In the self-attention sub-layer, the representations interact within modalities:
\begin{equation}
    H^{vs}=\text{ATTN}(H^{v}, H^{v}, H^{v}), ~ H^{ls}=\text{ATTN}(H^{l}, H^{l}, H^{l})
\end{equation}
In the cross-attention sub-layer, the representations interact across modalities to integrate cross-modal information into their representations:
\begin{equation}
    H^{vc}=\text{ATTN}(H^{vs}, H^{ls}, H^{ls}), ~ H^{lc}=\text{ATTN}(H^{ls}, H^{vs}, H^{vs})
\end{equation}
Finally, $H^{vc}$ and $H^{lc}$ are input to the feedforward sub-layer (i.e., an MLP) to obtain the multi-modal representations $Z^{v} = [z^{v}_{I}; z^{v}_{1}; z^{v}_{2}; ...; z^{v}_{N}]$ for vision and $Z^{l} = [z^{l}_{T}; z^{l}_{1}; z^{l}_{2}; ...; z^{l}_{M}; z^{l}_{SEP}]$ for language.

\subsection{Multi-modal Masked Autoencoders}
The idea of masked autoencoders has achieved great success in natural language processing (i.e., BERT) and recently in computer vision (i.e., MAE \cite{he2021mae}) as well.
However, in the general VLP area, existing studies \cite{dou2021meter,kim2021vilt} mainly recovered the original tokens of masked texts (denoted as masked language modeling (MLM)) and demonstrated that reconstructing the original signals of masked images (denoted as masked image modeling (MIM)) hurts the pre-training performance.
The reason is that vision and language have different characteristics, and appropriate designs are desired to make masked autoencoders work in such a multi-modal setting.
In detail, we develop three essential yet straightforward designs.

\noindent\textbf{Masking Strategy}~
Information density is different between vision and language.
Languages are information-dense messages created by humans, and thus predicting only a few held-out tokens can induce a sophisticated language understanding task.
On the contrary, images are spatial redundant.\footnote{A missing patch can be reconstructed easily from visible neighboring patches.}
As a result, we use random sampling with a much greater masking ratio for images (i.e., 75\%) than for texts (i.e., 15\%) to remove redundancy in images and enable the model to acquire valuable features from both images and texts.

\noindent\textbf{Representation Selection for Reconstruction}~
Images and texts are abstracted at different levels, with pixels of images having a lower semantic level than tokens of texts.
In our model, their representations are aggregated layer-by-layer in a hierarchical way.
Therefore, to make the final learned representations of images at a high semantic level, we instead adopt the intermediate outputs of the multi-modal fusion module (i.e., the visual outputs from the $k$-th Transformer layer denoted as $Z^{vk}$) to perform the low-level construction task (i.e., MIM).
For MLM, we still use the final output $Z^{l}$ for the prediction of tokens since predicting missing words requires richer semantic information.

\noindent\textbf{Decoder Designs}~
The vision and language decoders aim to map the representations $Z^{vk}$ and $Z^{l}$ back to their original input image and text, respectively.
After being encoded by the backbone model (i.e., the encoder), $Z^{vk}$ and $Z^{l}$ are represented at a high semantic level.
For the vision decoder, its output is in the pixel space, which is of a lower semantic level.
Therefore, a Transformer model is introduced as the decoder to map $Z^{vk}$ to lower semantic representations to perform the low-level reconstruction.
For the language decoder, its targets (i.e., words) are abstracted at a high level, and thus its design is trivial (i.e., an MLP).

Finally, the MIM loss is computed using the mean squared error (MSE) between the reconstructed and original images in the pixel space, and the MLM loss is computed as the negative log-likelihood loss for the masked tokens.\footnote{Note that MLM and MIM are performed in different forward procedures.}

\section{Experiments}
\subsection{Pre-Training Setup}
We conduct our experiments on two datasets, i.e., ROCO \cite{pelka2018roco} and MedICaT \cite{subramanian2020medicat}, where the former contains over 81,000 medical image-text pairs and the latter consists of over 217,000 medical images with their captions and inline textual references.
For ROCO, we adopt their official splits, and for MedICaT, we randomly sample 1,000 images for validation, 1,000 images for testing, and the remaining images are used for training.
For pre-training, we use the training set of ROCO and MedICaT to train models with the pre-training tasks presented in Section \ref{sec:approach} together with the common image-text matching task \cite{chen2020uniter} by default.

For the implementation, we adopt the architecture of CLIP-ViT-B \cite{radford2021clip} for the vision encoder and the architecture of RoBERTa-base \cite{liu2019roberta} for the language encoder.
For the multi-modal fusion module, we set the number of Transformer layers $N_{m}=6$ with the dimension of hidden states set to 768 and the number of heads set to 12.
For all pre-training experiments, the models are trained with AdamW optimizer \cite{loshchilov2018adamw} for 100,000 steps.
The learning rates for uni-modal encoders (i.e., the vision encoder and the language encoder) and the multi-modal fusion module are set to 1e-5 and 5e-5, respectively.
We set the warm-up ratio to 10\%, and use the linear learning rate scheduler after warm-up.
We use center-crop to resize each image into the size of 288$\times$288.

\begin{table}[t]
\centering
\caption{Comparisons of our proposed approach with previous studies on the test sets of three Med-VQA datasets with respect to the accuracy metric.}
\label{table:vqa}
\begin{tabular}{@{}lccccccc@{}}
\toprule
\multirow{2}{*}{Methods} & \multicolumn{3}{c}{VQA-RAD}                      & \multicolumn{3}{c}{SLACK}                        & VQA-2019       \\ \cmidrule(l){2-8} 
                                    & Open           & Closed         & Overall        & Open           & Closed         & Overall        & Overall         \\ \midrule
MFB \cite{yu2017mfb}                & 14.50          & 74.30          & 50.60          & 72.20          & 75.00          & 73.30          & -               \\
SAN \cite{yang2016san}              & 31.30          & 69.50          & 54.30          & 74.00          & 79.10          & 76.00          & -               \\
BAN \cite{kim2018ban}               & 37.40          & 72.10          & 58.30          & 74.60          & 79.10          & 76.30          & -               \\ \midrule
MEVF-SAN \cite{nguyen2019mevf}      & 49.20          & 73.90          & 64.10          & 75.30          & 78.40          & 76.50          & 68.90           \\
MEVF-BAN \cite{nguyen2019mevf}      & 49.20          & 77.20          & 66.10          & 77.80          & 79.80          & 78.60          & 77.86           \\
CPRD-BAN \cite{liu2021cprd}         & 52.50          & 77.90          & 67.80          & 79.50          & 83.40          & 81.10          & -                \\ \midrule
Ours                                & \textbf{67.23} & \textbf{83.46} & \textbf{77.01} & \textbf{80.31} & \textbf{87.82} & \textbf{83.25} & \textbf{79.87}   \\ \bottomrule
\end{tabular}
\end{table}
\subsection{Vision-and-Language Transfer Tasks}
We evaluate all models on three medical image-text understanding tasks (i.e., Med-VQA, medical image-text classification, and medical image-text retrieval).

\noindent\textbf{Medical Visual Question Answering}~
This task requires answering natural language questions about medical images.
We use the official dataset split of three publicly available datasets to train and evaluate the models: VQA-RAD \cite{Lau2018VQARAD}, SLAKE \cite{liu2021slake}, and VQA-2019 \cite{abacha2019medvqa}, where questions in VQA-RAD and SLAKE are both categorized into two types (i.e., closed-ended and opened-ended).

\noindent\textbf{Medical Image-Text Classification}~
This task aims to produce the label given an image-text pair.
We train and evaluate the models on MELINDA \cite{wu2021melinda}, a biomedical experiment method classification dataset, with its official split.

\noindent\textbf{Medical Image-Caption Retrieval}~
There are two subtasks for this task, where image-to-text (I2T) retrieval requires retrieving the most relevant texts from a large pool of texts given an image and vice versa for text-to-image (T2I) retrieval.
We train and evaluate the models on the official split of ROCO.

For the evaluation metrics, the models in Med-VQA and medical text-image classification tasks are evaluated w.r.t. accuracy, while those in the retrieval task are assessed using Recall@K (K=1, 5, 10).
We run each experiment three times with different random seeds and report the mean of its corresponding metric(s).

\subsection{Comparisons with the State-of-the-Art}
\begin{figure}[t]
\centering
\begin{minipage}{\textwidth}
\begin{minipage}[t]{0.34\textwidth}
\makeatletter\def\@captype{table}
\centering
\caption{Classification results on the MELINDA dataset.}
\vspace{10pt}
\begin{tabular}{@{}lc@{}}
\toprule
Methods                         & Test            \\ \midrule
ResNet-101 \cite{he2016resnet}  & 63.84           \\ \midrule
RoBERTa \cite{liu2019roberta}   & 74.60           \\ \midrule
NLF \cite{wu2021melinda}        & 76.60           \\
SAN \cite{yang2016san}          & 72.30           \\
Ours                            & \textbf{78.50} \\ \bottomrule
\end{tabular}
\label{table:cls}
\end{minipage}
\begin{minipage}[t]{0.62\textwidth}
\makeatletter\def\@captype{table}
\raggedright
\caption{Image-to-text and text-to-image retrieval results on the ROCO test set.}
\vspace{10pt}
\begin{tabular}{@{}lrrrrrr@{}}
\toprule
\multirow{2}{*}{Methods} & \multicolumn{3}{c}{T2I}                          & \multicolumn{3}{c}{I2T}                          \\ \cmidrule(l){2-7} 
                                             & R@1             & R@5            & R@10           & R@1            & R@5            & R@10           \\ \midrule
ViLT \cite{kim2021vilt}                      &  9.75           & 28.95          & 41.40          & 11.90          & 31.90          & 43.20          \\
METER \cite{dou2021meter}                    & 11.30           & 27.25          & 39.60          & 14.45          & 33.30          & 45.10          \\ \midrule
Ours (ZS)                                    & 19.05           & 47.75          & 61.35          & 19.10          & 45.60          & 61.20          \\
Ours (FT)                                    & \textbf{22.20}  & \textbf{52.50} & \textbf{66.65} & \textbf{22.90} & \textbf{51.05} & \textbf{65.80} \\ \bottomrule
\end{tabular}
\label{table:ret}
\end{minipage}
\end{minipage}
\end{figure}
The main experimental results on all downstream tasks are shown in Table \ref{table:vqa}, \ref{table:cls}, \ref{table:ret}.
Our proposed approach achieves state-of-the-art results on all datasets.
For Med-VQA, compared with the advanced approach CPRD-BAN, the proposed method outperforms it by 14.7\% and 5.5\% w.r.t accuracy on open-ended and closed-ended questions on VQA-RAD.
It also achieves 2.1\% and 2.0\% overall improvements on the SLACK and VQA-2019 datasets, respectively.
For medical image-text classification,
our method outperforms previous uni-modal and multi-modal methods under the non-continued pre-training setting, where it outperforms NLF by approximately 1.9\%.
For medical image-text retrieval, the proposed approach outperforms previous studies to a large extent in both the zero-shot (ZS) and fine-tuning (FT) settings.

\subsection{Quantitative Analysis}
\begin{figure}[t]
\centering
\begin{minipage}{\textwidth}
\centering
\begin{minipage}[t]{0.4\textwidth}
\makeatletter\def\@captype{table}
\centering
\caption{Ablation study on the VQA-RAD test set.}
\vspace{10pt}
\centering
\begin{tabular}{@{}ccccc@{}}
\toprule
MIM    & MLM    & Open           & Closed         & Overall        \\ \midrule
\xmark & \xmark & 24.67          & 80.78          & 58.48          \\
\cmark & \xmark & 22.41          & 79.17          & 56.56          \\
\xmark & \cmark & 67.04          & 81.99          & 76.05     \\
\cmark & \cmark & \textbf{67.23} & \textbf{83.46} & \textbf{77.01} \\ \bottomrule
\end{tabular}
\label{table:ablation}
\end{minipage}
\begin{minipage}[t]{0.55\textwidth}
\makeatletter\def\@captype{figure}
\centering
\vspace{10pt}
\includegraphics[width=0.98\textwidth]{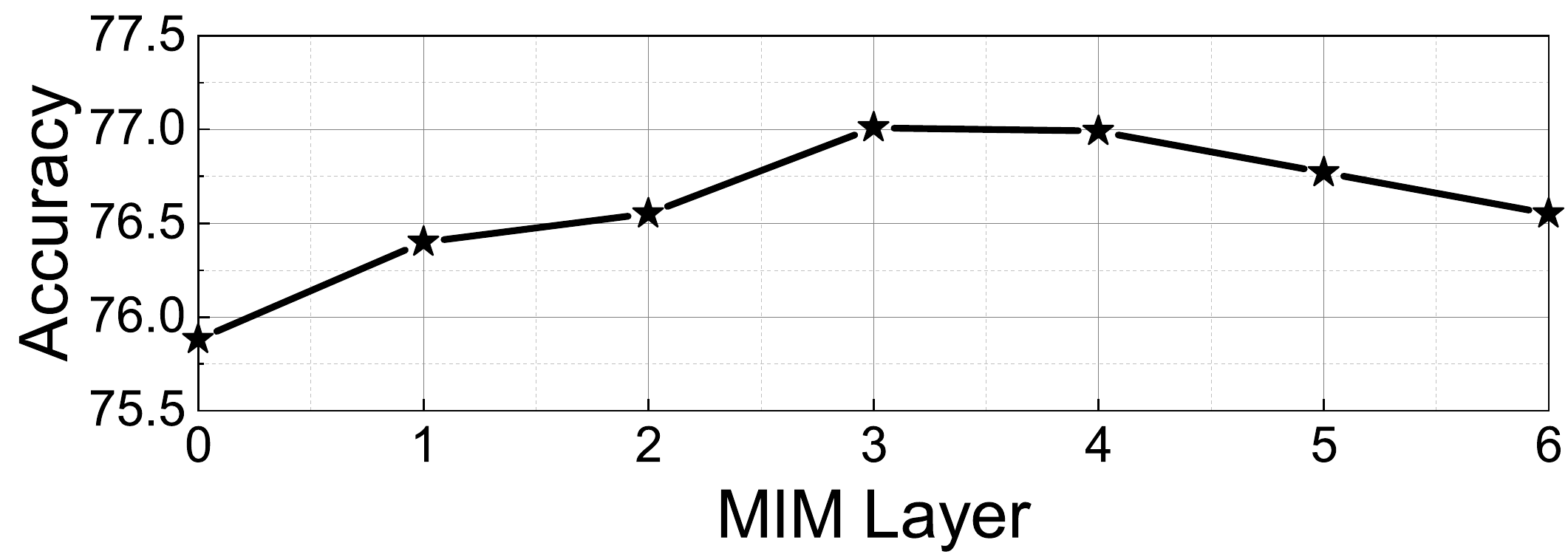}
\vspace{-2pt}
\caption{Results of selecting the representations from different layers to perform MIM.}
\label{fig:mim-layer}
\end{minipage}
\end{minipage}
\end{figure}
\noindent\textbf{Ablation Study}~
To demonstrate the effectiveness of different components of our approach, we perform an ablation study, with the results on VQA-RAD shown in Table \ref{table:ablation}.
There are several observations drawn from the results.
First, it is shown that only performing MIM in the pre-training can not bring an improvement when comparing between the 1st and 2nd rows.
Second, adopting MLM as one of the pre-training objectives (i.e., the 3rd and 4th rows) can achieve considerably better results than the case without MLM (i.e., the 1st and 2nd rows).
Third, the proposed M$^3$AE, with both MIM and MLM as its objectives, achieves the best result.
This owes to the fact that with the help of both MIM and MLM, the critical mappings between medical images and texts can be implicitly modeled, which promotes the learning of multi-modal representations.

\noindent\textbf{Effectiveness of different layers to perform MIM}~
To analyze the impacts of representations from different layers to perform MIM, we select the representations from layer 0 to 6 to pre-train our model, with the results shown in Fig. \ref{fig:mim-layer}.
There are two observations drawn from the results.
First, when using the representations from layer 0 (i.e., the visual features without any textual information), the method achieves the worst performance, verifying the importance of texts to perform MIM.
Second, it is observed that using the representations from the intermediate layer (i.e., layer 3) can achieve the best results.
This demonstrates that using the lower-level representations to perform MIM can help the model capture more hierarchical information in images and texts to keep the final learned image representations at a higher semantic level to promote representation learning.

\subsection{Qualitative Analysis}
\begin{figure*}[t]
\centering
\includegraphics[width=0.95\textwidth, trim=0 0 0 0]{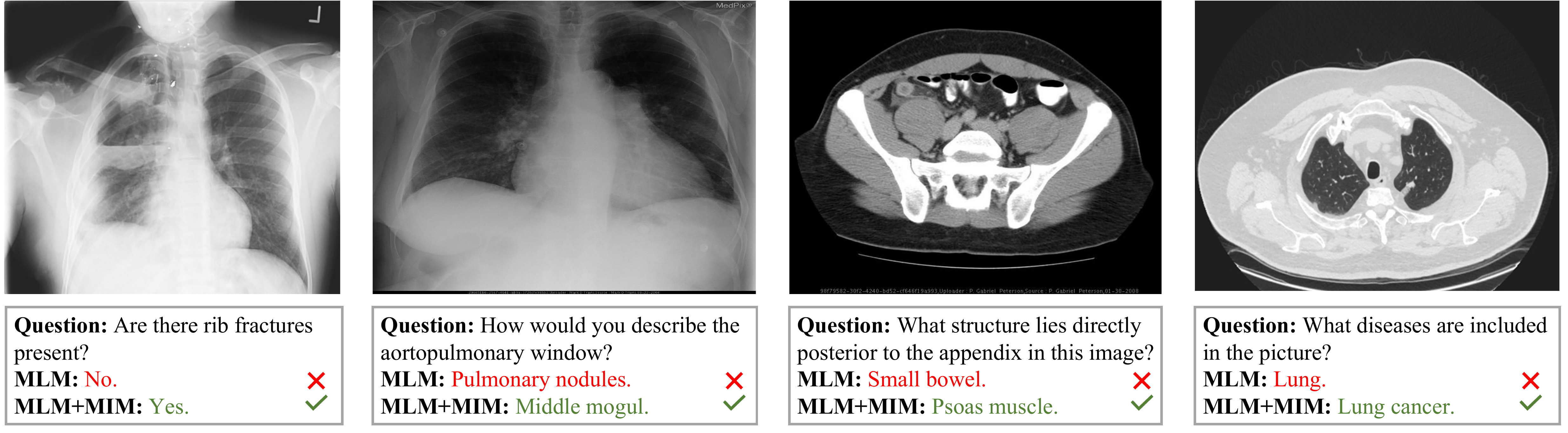}
\caption{Illustrations of four Med-VQA results from the models pre-trained with MLM and MLM+MIM on the VQA-RAD test set.}
\label{fig:case-study}
\end{figure*}
To further investigate the effectiveness of our method, we perform qualitative analysis on four Med-VQA cases on the VQA-RAD test set as shown in Fig. \ref{fig:case-study}.
In the 1st case, the MLM+MIM model answered an ``yes/no'' question correctly, and in the 2nd and 3rd cases, it answered the organ-related questions correctly while the MLM model did not.
Furthermore, in the last case, the MLM+MIM model answered the disease-related question correctly.
This demonstrates that pre-training with MLM+MIM might aid the model in learning more fine-grained mappings between images and texts.

\section{Conclusion}
In this paper, we believe that massive medical image-text data contains rich context and structural information, which is significant for medical vision-and-language understanding.
To this end, we propose an effective yet simple approach, i.e., multi-modal masked autoencoders, to perform pre-training on medical image-text pairs.
We develop three simple designs from the perspectives of masking ratios, representation selection for reconstruction, and decoder designs to make this simple approach work.
To evaluate our approach comprehensively, we construct a medical vision-and-language understanding benchmark, including three tasks.
Experimental results on various datasets demonstrate the superior performance of our approach, where state-of-the-art results are achieved.

\section*{Acknowledgement}
This work is supported in part by the Chinese Key-Area Research and Development Program of Guangdong Province (2020B0101350001), in part by the Guangdong Basic and Applied Basic Research Foundation (2020B1515020048), in part by the National Natural Science Foundation of China (61976250), in part by the Guangzhou Science and technology project (No.202102020633), and is also supported by the Guangdong Provincial Key Laboratory of Big Data Computing, The Chinese University of Hong Kong, Shenzhen.

\bibliographystyle{splncs04}
\bibliography{paper1841}
\end{document}